\documentclass[letterpaper, 10 pt, conference]{ieeeconf}  

\IEEEoverridecommandlockouts                              

\usepackage{cite}
\usepackage{amsmath,amssymb,amsfonts}

\usepackage{amsthm}
\usepackage{algorithmic}
\usepackage{graphicx}
\usepackage{epstopdf}
\usepackage{color}
\usepackage{textcomp}
\usepackage{dsfont}
\usepackage{verbatim}
\usepackage[linesnumbered,ruled]{algorithm2e}
\usepackage{hyperref}
\usepackage{tabularx}

\newcommand{\R}{\mathbb{R}} 
\newcommand{\figref}[1]{Fig. \ref{#1}}

\newtheorem{remark}{Remark}

\title{\LARGE \bf
Inverse Dynamics Control of Compliant Hybrid Zero Dynamic Walking
}

\author{Jenna Reher and Aaron D. Ames$^{1}$
    \thanks{*This research is supported under NSF Grant Numbers 1544332, 1724457, 1724464 and Disney Research LA.}
    \thanks{$^{1}$The authors are with the Department of Mechanical and Civil Engineering, California Institute of Technology, Pasadena, CA 91125
    {\tt\small \{jreher,ames\}@caltech.edu}}%
}

\begin{document}

\maketitle
\thispagestyle{empty}
\pagestyle{empty}

\begin{abstract}
We present a trajectory planning and control architecture for bipedal locomotion at a variety of speeds on a highly underactuated and compliant bipedal robot. A library of compliant walking trajectories are planned offline, and stored as compact arrays of polynomial coefficients for tracking online. The control implementation uses a floating-base inverse dynamics controller which generates dynamically consistent feedforward torques to realize walking using information obtained from the trajectory optimization. The effectiveness of the controller is demonstrated in simulation and on hardware for walking both indoors on flat terrain and over unplanned disturbances outdoors. Additionally, both the controller and optimization source code are made available on GitHub.
\end{abstract}

\section{INTRODUCTION}
The inclusion of model detail in walking controllers for underactuated bipedal locomotion can be used to accurately capture the underactuated dynamics of the system in a manner which also facilitates accurate tracking of planned motions. 
However, implementing model-based planning and control methods on physical systems is typically non-trivial due to the inherent model inaccuracy, dynamically changing contact constraints, and possibly conflicting objectives for the robot which naturally arise in locomotion.  
It is due to these challenges that bipedal robots which exhibit simultaneously robust, efficient, and agile motions are rare in practice \cite{reher2021annualreview}. 

A significant subset of the bipedal robotics literature mitigates the complexity of humanoids and bipeds by viewing walking as a problem wherein the real world dynamics are assumed to be governed by the evolution of a simpler system, such as a LIP models (Linear Inverted Pendulum \cite{2003Kajita, gong2020angular}), SLIP models (Spring Loaded Inverted Pendulum \cite{rezazadeh2015spring}), and the ZMP (Zero Moment Point \cite{vukobratovic2004zero}). 
These methods can reduce computational complexity for fast planning and experimental success. 
Despite its viability in practical implementation, this local representation of the system can limit the agility of behaviors and compromise energy efficiency.

\begin{figure}[t]
	\centering
	\includegraphics[width= 1\columnwidth]{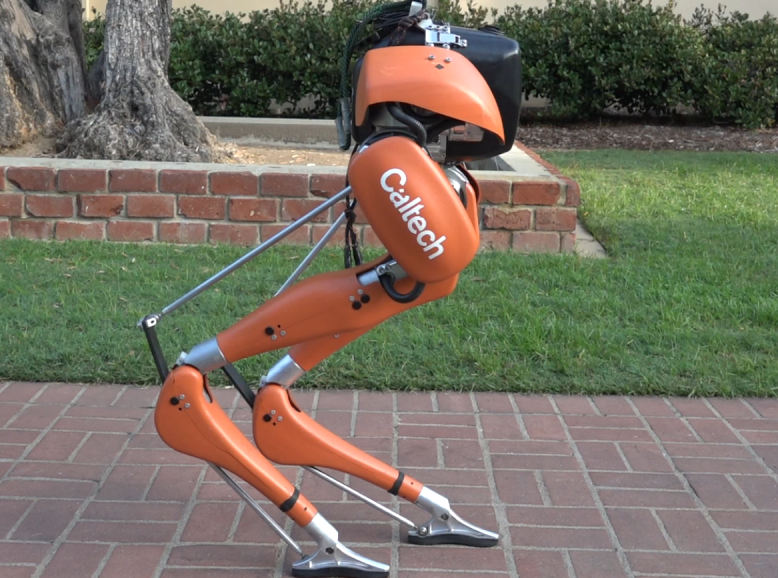}
	\caption{The bipedal robot Cassie at Caltech, used as the experimental platform for demonstration of the methods presented in this work.}
	\vspace{-6mm}
	\label{fig:cover}
\end{figure}

One area of model-based planning and control which is particularly difficult to directly address is passive compliance in locomotion. 
Some of the earliest inclusions of compliant hardware on bipedal robots was with spring flamingo and spring turkey \cite{hunter1991comparative}, with more recent examples being MABEL \cite{park2011identification} and ATRIAS \cite{rezazadeh2015spring}.
One of the latest robots to exhibit compliant leg structures is the Cassie biped (shown in \figref{fig:cover}), which is the experimental platform that we consider in this work. 
From a mathematical standpoint, compliance increases the degree of underactuation which makes finding walking behaviors that satisfy stability constraints more difficult. With regard to implementation compliance also increases numerical stiffness and model uncertainty.  We attempt to address these challenges through two contributions. 

The first contribution of this paper is a motion library of walking behaviors for Cassie that leverage its full-body dynamics including its compliance. 
To generate this library, we utilize the framework of Hybrid Zero Dynamics (HZD) \cite{westervelt2018feedback} which has demonstrated success in developing controllers for highly underactuated walking behaviors. 
The HZD framework \cite{westervelt2018feedback} has demonstrated success in developing controllers for highly underactuated walking behaviors. 
In the context of robotic implementations, HZD has enabled a wide variety of dynamic behaviors such as underactuated humanoid walking \cite{reher2020algorithmic}, compliant running \cite{sreenath2013embedding}, and 3D bipedal walking with point-feet \cite{ramezani2014performance}. 
Directly related to this work, motion libraries on sagittal motions under the assumption of sufficient rigidity has been successfully realized on Cassie \cite{gong2018feedback}, but this work ignored the robot's inherent compliance.  Tangentially, singular walking behaviors which consider passive compliance were realized in \cite{reher2019cassie}.  This work, therefore, is the first to combine motion libraries that consider the full-body dynamics of Cassie together with compliance. 

The second contribution of this paper is the synthesis of controllers able to realize motion libraries in a manner that exploits the compliance for which they were generated.  
One of the limiting assumptions which can affect HZD behaviors is that their formal stability guarantees are often tied to exponential tracking of motion objectives, or outputs. 
This can lead to the use of high-gain PD feedback controllers when applied to hardware to obtain periodic stability. 
Instead, it would be more desirable to achieve tracking with some inherent control compliance to perturbations and unknown terrain. 
In this direction, model-based control in the form of inverse dynamics \cite{aghili2005unified, koolen2016design}, control Lyapunov functions \cite{ames2014rapidly}, or some combination of the two \cite{reher2020inverse} may offer more desirable control properties. 
In this work, we consider an analytical solution to the floating-base inverse dynamics problem, extending the work in \cite{mistry2010decompprojection}. 
The end result is a model-based controller that is 
demonstrated experimentally on Cassie, with the result being stable and robust walking.

The paper is structured as follows: Section \ref{sec:robotmodel} gives an overview of the Cassie robotic model and the hybrid domain structure which is prescribed for walking. Section \ref{sec:opt} details the optimization problem which is used to produce a library of compliant walking trajectories. Section \ref{sec:implementation} introduces a orthogonal decomposition based inverse dynamics controller, which uses the optimized accelerations and robot model to obtain a dynamically consistent feedforward term for the planned walking gaits. Walking on hardware and in simulation is then demonstrated, validating that the planned trajectories capture the passive dynamics of the robot.

\section{Robotic Model}
\label{sec:robotmodel}
The Cassie biped is an approximately one meter tall walking robot designed and manufactured by Agility Robotics\footnote{http://www.agilityrobotics.com/}. 
The sensing on the robot is entirely proprioceptive and includes a 9-DOF IMU, torque sensing, and high-resolution absolute encoders to measure both compliance and actuated joints. The physical model of Cassie is complex, with 22 degrees of freedom, 4 of which are fiberglass leaf springs. 
There are $10$ BLDC motors which control the joints through $8$ low-friction cycloidal gearboxes and $2$ harmonic gearboxes at the feet. 
The compliance in the model is contained to a 4-bar linkage, through which the knee motors effectively drive the leg length. 
While the feet of the robot are actuated, and offer some small control authority in the sagittal direction, contact is a reduced to a point in the coronal plane. 

\subsection{Floating Base Model}
\label{sec:config}
\begin{figure}[t]
	\centering
	\vspace{2mm}
	\includegraphics[width= 1\columnwidth]{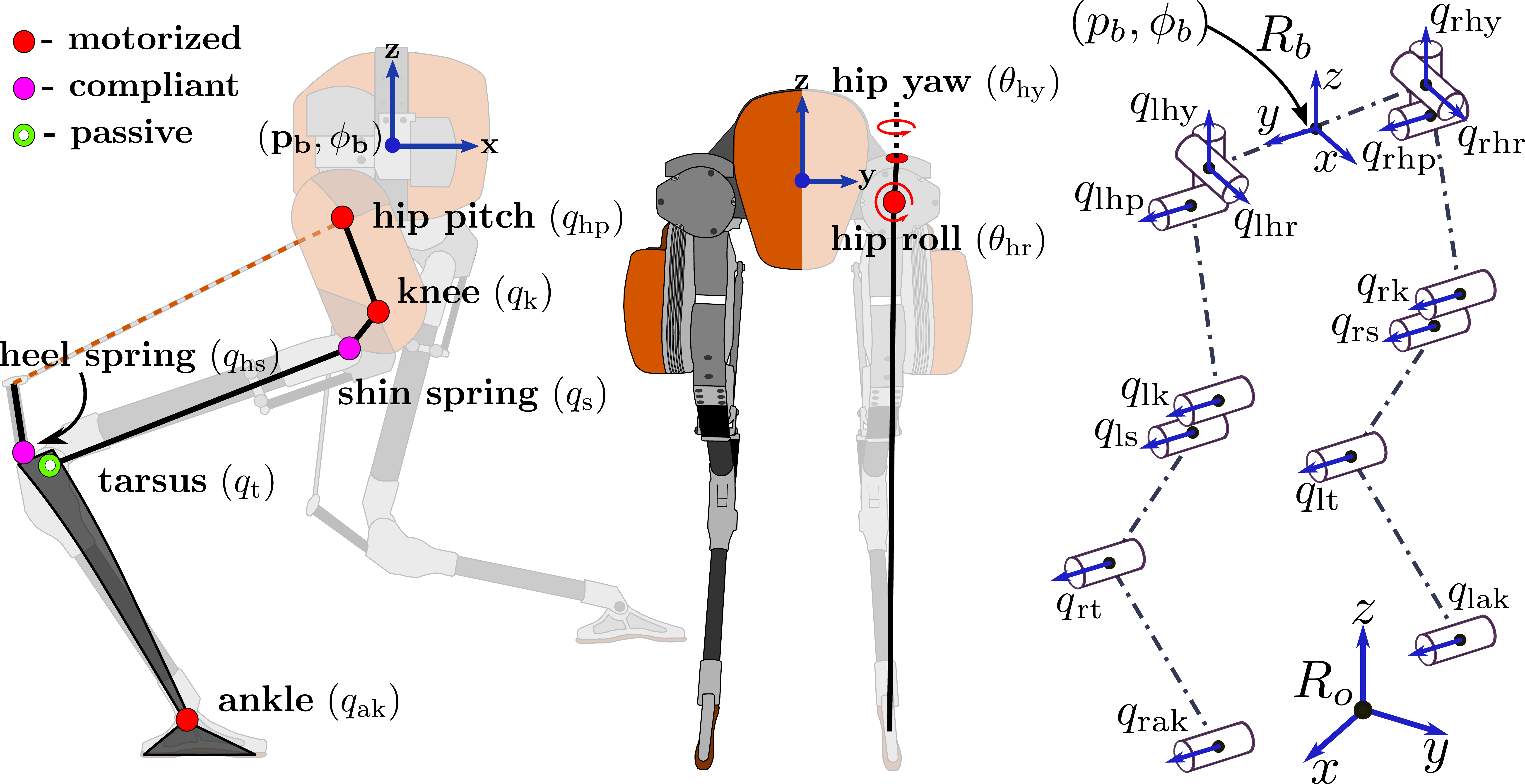}
	\caption{The configuration coordinates of the Cassie robot: with the side and front views highlighting the compliant mechanism and the general morphology of the robot.
    }
	\label{fig:cassie_configuration}
	\vspace{-5mm}
\end{figure}

The robot itself is modeled as a tree structure composed of rigid links. As legged locomotion inherently involves intermittent sequences of rigid contacts with the environment, it is common practice to construct a floating-base Lagrangian model of the system and then add constraint forces via D'Alembert's principle \cite{Grizzle2014Models}. We define the configuration space of the bipedal robot Cassie as $\mathcal{Q} \subset\R^n$, where $n$ is the DOF (degrees of freedom) without considering any holonomic constraints that result in DOF reduction. A visualization of the prescribed coordinates is shown in \figref{fig:cassie_configuration}, and an URDF of the robotic model detailed in this work is provided online as a part of a software package \cite{papergithub}. Let 
\begin{align*}
    q = (p_b, \phi_b, q_l) \in \mathcal{Q} := \R^3\times SO(3)\times \mathcal{Q}_l,
\end{align*}
where $p_b$ is the global Cartesian position of the body fixed frame attached to the base linkage (the pelvis), $\phi_b$ is its global orientation, and $q_l\in\mathcal{Q}_l\subset\R^{n_l}$ are the local coordinates representing the joint angles between links. Further, the state space $\mathcal{X}=T\mathcal{Q}\subset \R^{2n}$ has coordinates $x=(q^\mathsf{T},\dot q^\mathsf{T})^\mathsf{T}$. 
The local coordinates are denoted as $\displaystyle q_l^\mathsf{T} = \big( q^{\mathrm{L}}, q^{\mathrm{R}} \big)$,
where the superscript $\mathrm{L}/\mathrm{R}$ denotes the left/right leg, and 
\begin{align*}
q^{i\in \{\mathrm{L,R}\}} = \big(
    \theta_\mathrm{hr}^{i}, \theta_\mathrm{hp}^{i}, \theta_\mathrm{hy}^{i},  \theta_\mathrm{k}^{i},
    \theta_\mathrm{s}^{i},  \theta_\mathrm{t}^{i},  \theta_\mathrm{hs}^{i},  \theta_\mathrm{a}^{i}
    \big),
\end{align*}
represents the joint of \textit{hip roll, hip pitch, hip yaw, knee, shin, tarsus, heel spring and ankle} accordingly. 
Among these joints, $\theta_\mathrm{hr}, \theta_\mathrm{hp}, \theta_\mathrm{hy}, \theta_\mathrm{k}, \theta_\mathrm{a}$ are actuated by BLDC motors with associated control inputs $u\in\mathcal{U}\subset\R^m$ with $m=10$ as previously described. The joints $\theta_\mathrm{hs}, \theta_\mathrm{s}$ are directly driven by leaf springs, which are treated as rigid links with rotational springs of stiffness $2300$ and $2000$ Nm/rad respectively at the pivot. There is one completely passive joint $\theta_\mathrm{t}$ per leg, and an unconstrained model of the robot with $22$ DOF. 

The floating-base dynamic model can be expressed in the form of Lagrange:
\begin{align}
    \label{eq:eom}
    D(q) \ddot{q} + H(q,\dot{q}) = B u + J (q)^\mathsf{T} \lambda + B_s \tau_s 
\end{align}
where $D(q)$ is the inertia matrix, $h(q,\dot q)$ contains the Coriolis and gravity terms, $B$ is the actuation matrix with gear reduction as its entries, and the Jacobian matrix of the holonomic constraints is $J(q)=\partial \frac{\eta}{\partial q}(q)$ with its corresponding constraint wrenches $\lambda\in\R^{m_{\eta}}$. Note that we introduced the spring forces as external forces $\tau_s = k_s q + k_b \dot q$ with $k_s$, $k_b$ the stiffness and damping coefficients, which are distributed according to the associated spring selection matrix $B_s$. 

\subsection{Holonomic Constraints} 
Two types of holonomic constraints are commonly considered for legged robotic systems, external contact constraints depending on the current configuration of the robot and it's interactions with the world, and internal kinematic constraints resulting from the robot geometry. Both types of constraints are enforced in the same manner, by prescribing a closure equation, $\eta(q) = \mathrm{constant}$, which represents a positional constraint on the robot. This expression then yields a constraint on the accelerations when differentiated twice:
\begin{align}
\label{eq:hol_accel}
    J(q) \ddot{q} + \frac{\partial}{\partial q} \left( \frac{\partial J(q)}{\partial q} \dot{q} \right) \dot{q} = 0.
\end{align}
The enforcement of this constraint gives rise to the corresponding force terms, $\lambda$, in the equations of motion \eqref{eq:eom}.

\vspace{0.2cm}
\noindent\textbf{Contact Constraints.} 
In this work, ground contact is assumed to be enforced through a holonomic constraint on a stance foot's position and orientation, $\eta_{st}(q)$, which is rendered constant while the foot remains on the ground. Because the width of the feet on Cassie is negligible we enforce contact as a line, or as two collinear points of contact at the toe and heel. This gives rise to a ground contact constraint for each foot with five components:
\begin{align}
\label{eq:hol_contact}
    \eta_{st}(q)^{\mathsf{T}} := \left[ p_{st}^x, p_{st}^y, p_{st}^z, \phi_{st}^y, \phi_{st}^z \right]^\mathsf{T},
\end{align}
where the first three components are the Cartesian position of the center of the foot and the last two correspond to the the foot pitch and yaw, shown in \figref{fig:direct}. Because $\lambda_{st}^z$ is a normal force, it is unilateral. Additionally, the tangential forces $\lambda_{st}^x$, $\lambda_{st}^y$ are required to satisfy friction models to remain feasible. Ideally, a classical \textit{Amontons-Coulomb model} of (dry) friction is used to avoid slippage and is represented as a friction cone. For a friction coefficient $\mu$ and a surface normal, the space of valid reaction forces is then:

\vspace{-2mm}
{\small
\begin{align*}
    \mathcal{C} = \left\{ \left. ( \lambda_{st}^x, \lambda_{st}^y, \lambda_{st}^z ) \in \R^3 \right| \lambda_{st}^z \geq 0; \sqrt{(\lambda_{st}^x)^2 + (\lambda_{st}^y)^2} \leq \mu \lambda_{st}^z \right\}. 
\end{align*}}
Additionally, the moment associated with the foot pitch $\lambda_{st}^{my}$ can produce rotation of the foot over the forward edge if it is too large. It has been shown that due to the unilateral nature of contact this moment is limited by:
\begin{align}
      -\frac{l}{2} \lambda_{st}^z <  &\lambda_{st}^{my} < \frac{l}{2} \lambda_{st}^z,
\end{align}
where $l$ is the length of the foot from heel to toe.

\vspace{0.2cm}
\noindent\textbf{Loop Constraints.} 
It is common practice to model robotic manipulators in branched tree structures. 
However, on Cassie, a \textit{compliant multi-link mechanism} is used to transfer power from higher to lower limbs without allocating the actuators' major weight onto the lower limbs, and effectively acts as a pair of springy legs \cite{reher2019cassie}. 
When the mechanism has a kinematic loop, this is often managed by cutting the loop at one of the joints and enforcing a holonomic constraint at the connection to form the closed-chain manipulator. 
In the Cassie leg the heel spring is attached to the rear of the tarsus linkage, with its end constrained via a pushrod affixed to the hip pitch linkage. For this work, we assume that pushrod attachment is a virtual holonomic distance constraint applied between the hip and heel spring connectors as:
\begin{align}
\label{eq:hol_ach}
    \eta_{ach}(q_l) := d(q_l) - 0.5012 = 0, 
\end{align}
where the attachment distance $d(q_l)\in\mathbb{R}$ is obtained via the forward kinematics between connectors at the hip and heel spring (see \figref{fig:cassie_configuration}). We also assume that when a leg is in swing that the springs on that leg are a rigidly fixed:
\begin{align}
    \eta_{sw}(q)^\mathsf{T} := [q_s, q_{hs}]^\mathsf{T} = 0. \label{eq:swing_rigid_constraint}
\end{align}
This simplifies both the optimization and control implementations \cite{sreenath2013embedding}, and makes the dynamics less numerically stiff.

\subsection{Hybrid Locomotion Model}
A bipedal walking gait consists of one or more different continuous
phases followed by discrete events that transition from one
phase to another, motivating the use of a hybrid system formulation 
with a specific ordering of phases. This is traditionally described as a
walking cycle, which is a directed cycle with a sequence of continuous
domains (continuous dynamics) and edges (changes in contact conditions).

\begin{figure}
\centering
\vspace{2mm}
	\includegraphics[width= 1\columnwidth]{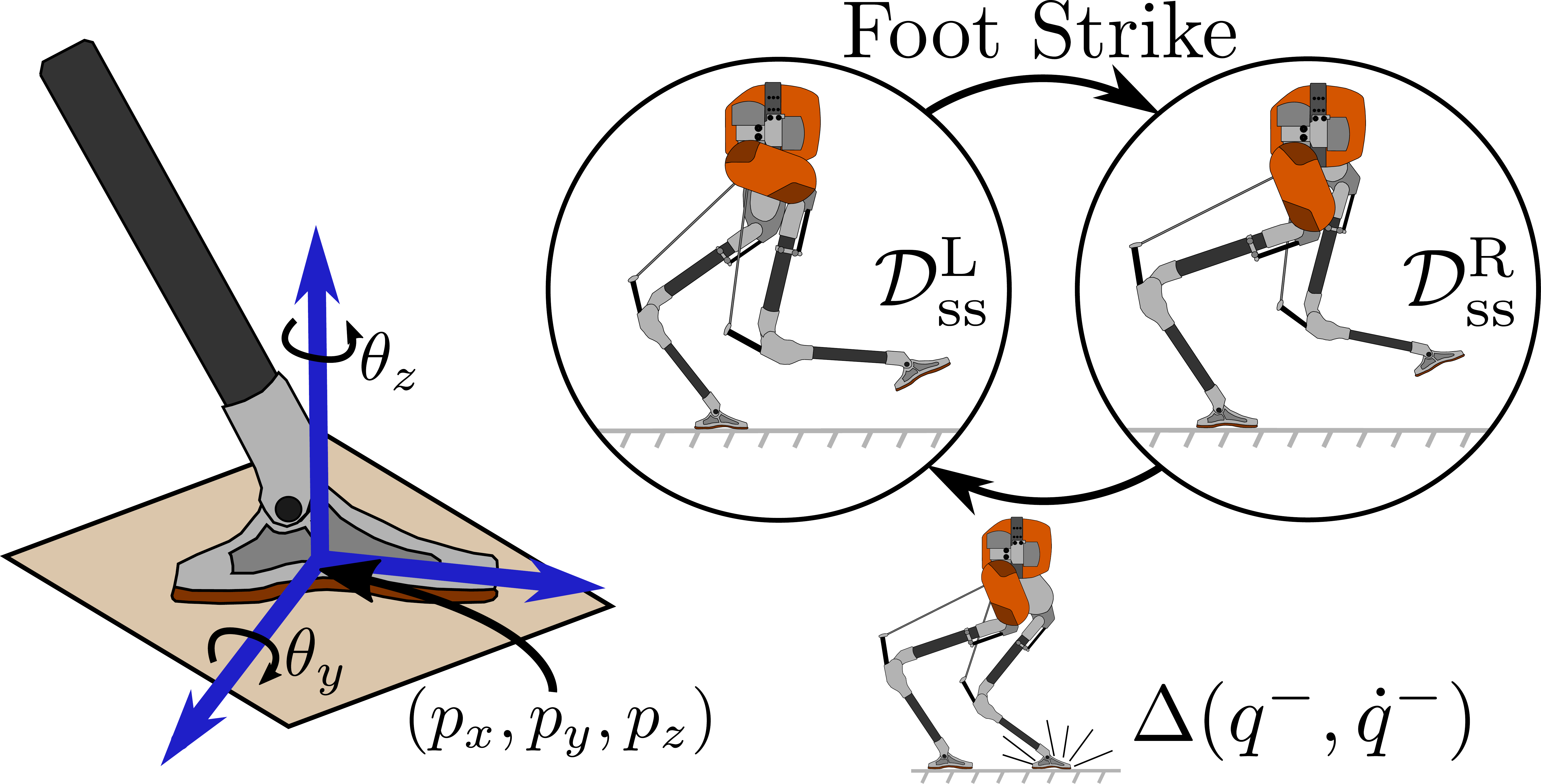}
	\caption{Shown on the left is the contact geometry of the robot's foot, along with the constrained coordinates and associated contact frame. The directed graph of walking used in this paper is shown on the right, where we view walking on Cassie as consisting of two single-support domains with a compliant stance leg, and rigidly stiff swing leg. }
	\label{fig:direct}
	\vspace{-5mm}
\end{figure}

\begin{figure*}[t]
	\centering
	\vspace{2mm}
	\includegraphics[width= 1\textwidth]{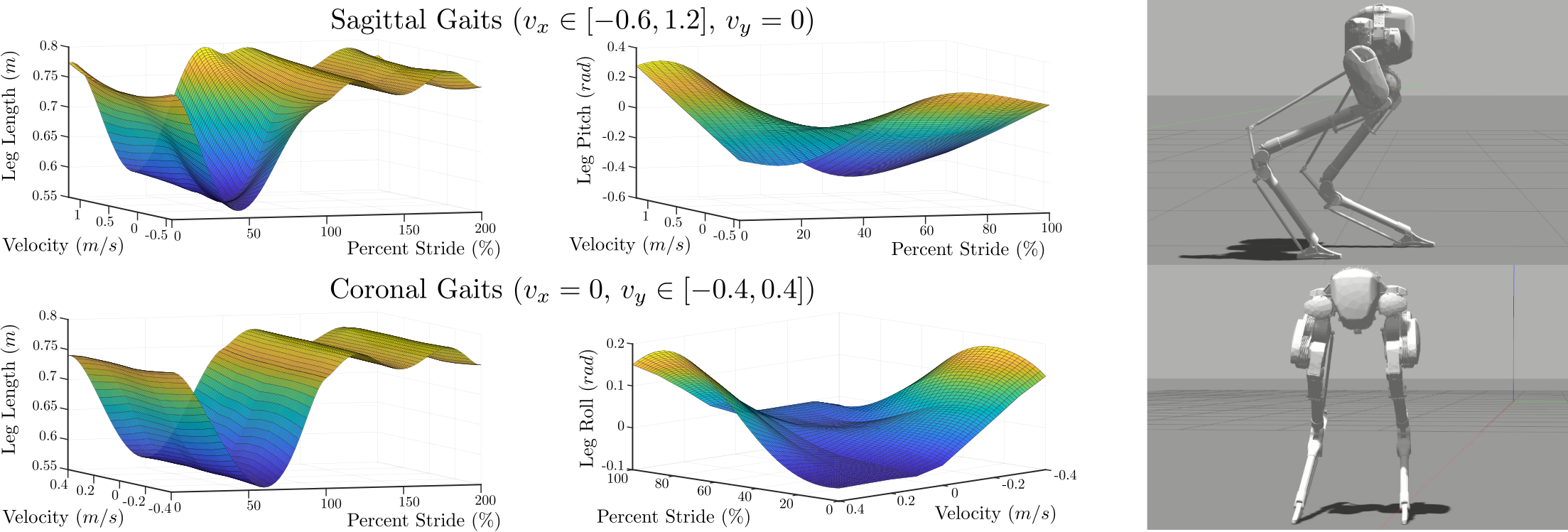}
	\caption{Contour plots of the swing leg length, leg angle, and leg roll outputs over the library speeds in the sagittal and coronal directions, showing the forward and reverse sweep of the leg as it tracks the motions. Also shown is the corresponding motion executed in a Gazebo simulation.}
	\label{fig:opt_output_continuum_feedforward}
	\vspace{-4mm}
\end{figure*}

In this work, we structure the dynamics of walking on Cassie in a hybrid fashion. The walking consists of two single support domains, $\mathcal{D}_\mathrm{SS}^{\{ \mathrm{L, R} \}}$, associated with stance on the respective left (L) or right (R) foot, with an instantaneous double-support (the swing leg lifts immediately on impact). An associated directed graph can be specified for the system which is depicted in \figref{fig:direct}. There is an additional assumption that the walking is symmetric for both left and right stance when the robot's average lateral speed over a step is zero, while the walking is allowed to be asymmetric otherwise. A transition from one single support domain to another occurs when the normal reaction force on the non-stance foot crosses zero. Therefore, the domain and guard are given by:
\begin{align*}
    \mathcal{D}_\mathrm{SS}^{\{ \mathrm{L, R} \}} &= \{ (q, \dot{q}, u) :  p_{swf}^z(q) \geq 0, \lambda_{nsf}^z(q,\dot{q},u) = 0\}, \\
    S_{\{ \text{L}\rightarrow\text{R, } \text{R}\rightarrow\text{L} \}} &= \{ (q,\dot{q}) :  p_{swf}^z(q) = 0, \dot{p}_{swf}^z(q,\dot{q}) < 0 \},
\end{align*}
where $\lambda_{swf}^z(q,\dot{q},u)$ is the vertical ground reaction force of the swing foot and $p_{swf}^z(q)$ is the vertical position of the center of the swing foot from the ground. 
Impact occurs when the swing foot touches the ground, modeled here as an inelastic contact between two rigid bodies. 
In this contact model, the configurations of the robot are invariant through the impact and velocities will instantaneously change due to the introduction of new holonomic constraints \cite{Grizzle2014Models, Hurmuzlu1994Rigid}. The associated reset map, $\Delta$, is given as:
\begin{align}
  \label{eq:reset-map}
  \Delta := \left[\begin{array}{c}
      q^+ \\
      \dot{q}^+
    \end{array}
  \right] = \left[\begin{array}{c}
      \mathcal{R} q^- \\
      \mathcal{R} \Delta^{\dot{q}}(q^-) \dot{q}^-
    \end{array}
  \right],
\end{align}
where $\mathcal{R}$ is a relabeling matrix, $q^-$ and $q^+$ denote the pre and post-impact configuration, and $\Delta^{\dot{q}}(q)$ is obtained from the plastic impact equation:
\begin{align}
\label{eq:vel_impact}
  \Delta^{\dot{q}}(q^-) = I - D^{-1} J^\mathsf{T}(J
  D^{-1} J^\mathsf{T})^{-1} J.
\end{align}

\section{Trajectory Planning for Walking}
\label{sec:opt}
This section details the trajectory optimization approach used to design a collection of trajectories which can be implemented to obtain compliant walking behaviors on the physical system in both the sagittal and coronal directions.

\subsection{Virtual Constraints}
Analogous to holonomic constraints, virtual constraints are defined as a set of functions that regulate the motion of the robot with a desired behavior \cite{westervelt2018feedback}. The term ``virtual'' comes from the fact that these constraints are enforced through feedback controllers instead of through physical constraints. The primary idea is to design a controller $u(x, \alpha)$ to regulate:
\begin{eqnarray}
\label{eq:outputs}
y(q,t) := y^a(q) - y^d(\alpha,t),
\end{eqnarray}
for which a behavior can be encoded through the desired outputs, $y^d(\alpha,t)$. A $6$th-order B\'ezier polynomial chosen for the desired outputs, for which $\alpha$ is a matrix of real coefficients that parameterize the curve. In this work, we consider nine outputs of vector relative degree $2$: 
\begin{align*}
\label{eq:actual_outputs}
    y^a(q) := 
    \left.
    \begin{bmatrix} 
        \phi_x, \ \phi_y\\
        \theta_{hy}^{st}, \ \theta_{hy}^{sw}\\
        \theta_{hr}^{sw}\\
        ||\psi^{sw}||_{2}\\ 
        ||\psi^{st}||_{2}\\ 
        \textrm{\footnotesize{atan2}}\big(\psi^{sw}_x,\psi^{sw}_z\big)\\
        \phi^y(q)
    \end{bmatrix} 
    \right\vert_{\begin{matrix} q_{\mathrm{s}=0} \\ q_{\mathrm{hs}=0} \end{matrix}}
    \begin{pmatrix}
        \textrm{\footnotesize{pelvis roll/pitch}}\\
        \textrm{\footnotesize{stance/swing hip yaw}}\\
        \textrm{\footnotesize{swing hip roll}}\\
        \textrm{\footnotesize{swing leg length}}\\
        \textrm{\footnotesize{stance leg length}}\\
        \textrm{\footnotesize{swing leg pitch}}\\
        \textrm{\footnotesize{swing foot pitch}}\end{pmatrix} 
\end{align*}
where $\phi^y(\theta_{\textrm{tp}})$ is the swing foot Cartesian pitch and,
\begin{equation}
    \psi^{st/sw}(q) = p_{hp}^{st/sw}(q) - p_{tp}^{st/sw}(q),
\end{equation}
is the expression for the distance between the hip pitch and ankle pitch joints from the forward kinematics. Because the foot geometry is quite small, we leave the stance foot passive. It should be noted that we are controlling the undeflected leg length and angle of the legs by zeroing the spring deflections. By formulating the outputs in this way, the passive dynamics of the system will contain the additional dynamics associated with the compliant elements \cite{sreenath2013embedding}. As a practical matter, this is also important as directly controlling the compliance in the leg is a significantly more difficult problem to achieve.

\subsection{Hybrid Zero Dynamics} 
The combined mechanical model \eqref{eq:eom} and \eqref{eq:hol_accel} can be expressed in state variable form as, 
\begin{align}
\label{eq:eom_nonlinear}
    \dot{x} = f(x) + g(x) u.
\end{align}
Additionally, because all outputs specified for the system have vector relative degree two, we assume the existence of a feedback control law $u^*(u,t)$ that drives $y \to 0$ exponentially, resulting in a closed loop system: 
\begin{align}
\label{eq:eom_nonlinearcl}
    \dot{x} = f_{\mathrm{cl}}(x) = f(x) + g(x) u^*(x).
\end{align}
Additionally, by driving the outputs to zero the controller $u^*$ renders the 
\textit{zero dynamics manifold}: 
\begin{align}
\label{eq:zerodyn}
    \mathbb{Z} = \{ (q,\dot{q}) \in \mathcal{D} ~  | ~ y(q,t) = 0 ,~   L_{f_{\mathrm{cl}}} y (q,\dot{q},t) = 0 \}.
\end{align}
forward invariant and attractive. Thus, the continuous dynamics \eqref{eq:eom_nonlinearcl} will then evolve on $\mathbb{Z}$. However, because \eqref{eq:zerodyn} has been designed without taking into account the hybrid transition maps \eqref{eq:reset-map}, it will not be impact invariant. In order to enforce impact invariance, the B\'ezier polynomials for the desired outputs can be shaped through the parameters $\alpha$. This can be interpreted as the condition:
\begin{align}\label{eq:resetmapv}
 \Delta ( \mathbb{Z} \cap S ) \subset \mathbb{Z},
\end{align}
and will be imposed as a constraint on the states through the impact \eqref{eq:reset-map}. When \eqref{eq:resetmapv} is satisfied, we say that the system lies on the \textit{hybrid zero dynamics} (HZD) manifold.

\subsection{Gait Library Optimization} 
The desired evolution of the outputs \eqref{eq:outputs} must now be designed such that we can achieve locomotion, while also satisfying both the physical limitations of the hardware and other conditions such as \eqref{eq:resetmapv}. A nonlinear trajectory optimization problem is formed to solve this problem. Similar to \cite{xie2020learning,gong2018feedback}, we would like to design a variety of walking speeds for which the robot can operate. To accomplish this, a library of walking gaits at sagittal speeds of $v_x\in [ -0.6,1.2 ]$ m/s and coronal speeds of $v_y \in [-0.4, 0.4]$ m/s are generated in a grid of $0.1$ m/s intervals. The impact-to-impact duration of each step is fixed at $0.4$ seconds for ease of implementation.

\begin{table}[b]
\centering
\caption{Optimization constraints and parameters}
\label{table:optimization_constraints}
\def\arraystretch{1.2} 
    \begin{tabular}{ | m{13.7em} m{2.3cm} m{0.40cm} | } 
        \hline
        Step duration  &  $=0.4$ & sec \\ 
        \hline
        Average step velocity, $\bar{v}_{x,y}$ & $=v_{x,y}$ & m/s \\
        \hline
        Pelvis height, $p_z$  &  $\geq 0.80$ & m \\ 
        \hline
        Mid-step foot clearance, $p_{nsf}^z$  &  $\geq 0.14$ & m \\ 
        \hline
        Vertical impact velocity, $\dot{p}_{sw}^z$  &  $\in(-0.40, -0.10)$ & m/s \\
        \hline
        Step width, ${p}_{lf}^y - {p}_{lf}^y$  &  $\in(0.14, 0.35)$ & m \\ 
        \hline
        Swing foot pitch, $\phi^y(q)$ & $=0$ & rad \\
        \hline
        Friction cone, $\mu$    &  $< 0.6$ &  \\ 
        \hline
    \end{tabular}
\end{table}

Each hybrid optimization was performed over the two domains, $\mathcal{D}_\mathrm{SS}^{\{ \mathrm{L, R} \}}$, with a constraint imposed such that when the discrete impact \eqref{eq:reset-map} is applied to the terminal state that it satisfies the hybrid invariance condition \eqref{eq:resetmapv}. It is critical that the motions respect the limitations of the physical system. In order to address this constraints for the friction cone, 
actuator limits, and joint limits are imposed. While these constraints alone ensure invariance and satisfaction of the physical constraints, additional constraints on the behavior were tuned for implementation on hardware, such as the swing foot velocity, impact foot configuration, step symmetry when $v_y=0$, and others outlined in Table \ref{table:optimization_constraints}. 

In order to minimize torque and to center the floating base coordinate movement around the origin, the following cost function was minimized:
\begin{align}
    \mathcal{J}(\mathbf{w}) &:= \int_{t=0}^{tf} \left( c_u |u|^2 + c_\phi |\phi_b|^2 \right) dt,
    \label{eq:cost}
\end{align}
where $\mathbf{w}\in\mathbb{R}^{N_w}$ with $N_w$ being the total number of optimization variables and $c_{\square}$ are weights applied to the respective terms. For the library presented in this work the weights used were $c_u = 0.0001$ and $c_\phi = (20,1,30)$.

\begin{figure}[t]
	\centering
	\vspace{1mm}
	\includegraphics[width= 1\columnwidth]{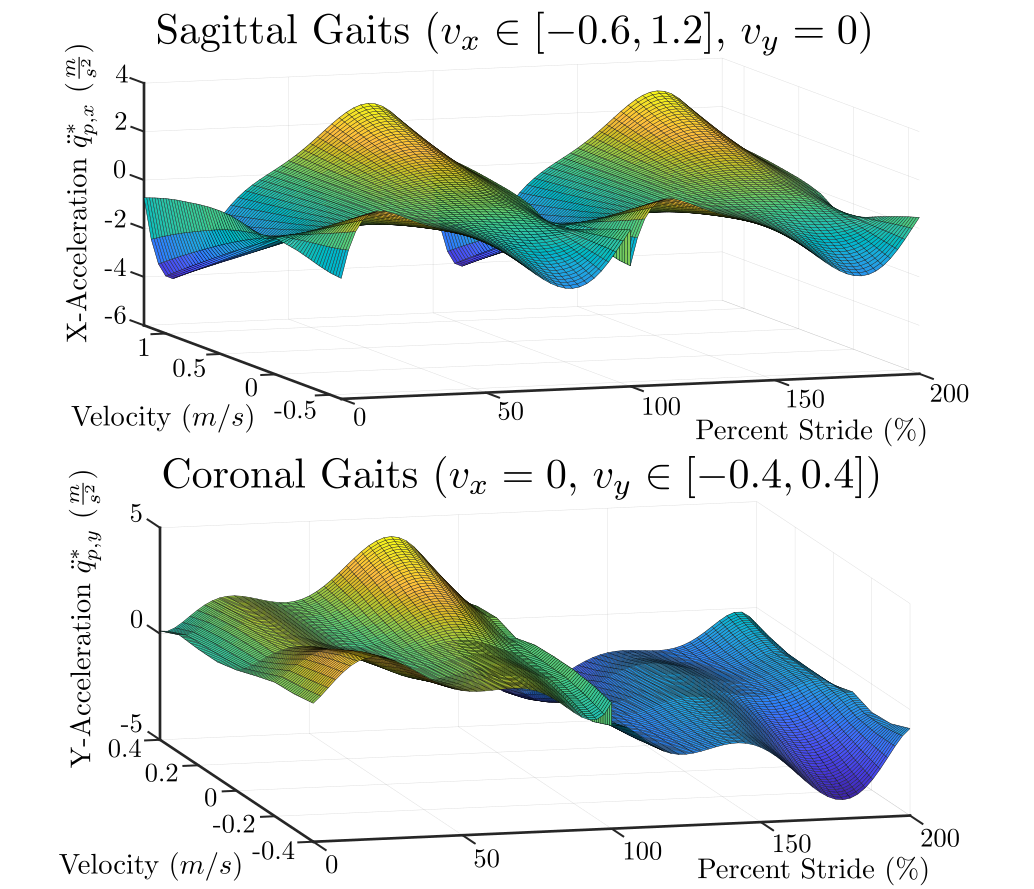}
	\caption{The contours of the floating base $x$ and $y$ accelerations which are obtained from the trajectory optimization problem. These contribute to the control action of a feedforward controller which is dynamically consistent with the contact constraints of the underactuated robotic walking.}
	\label{fig:ddq_plots}
	\vspace{-6mm}
\end{figure}

An optimization package, FROST \cite{hereid2017frost}, was used to transcribe the constraints and cost of each of the $171$ gaits into a nonlinear programming (NLP) problem that can be solved by a standard optimization solver such as IPOPT \cite{biegler2009large}:
\begin{align}
    \label{eq:opteqs}
	\mathbf{w}^* = \underset{\mathbf{w}}{\mathrm{argmin}} &\hspace{3mm} \mathcal{J}(\mathbf{w}) \\
	\mathrm{s.t.} 		&\hspace{3mm}  \text{Closed\ loop\ dynamics: Eq.\eqref{eq:eom_nonlinearcl}}  \notag \\
				  		&\hspace{3mm}  \text{HZD\ condition: Eq. \eqref{eq:resetmapv}}  \notag\\
				  		&\hspace{3mm}  \text{Physical feasibility: Table \ref{table:optimization_constraints}}    \notag
\end{align}
Each optimization was then solved through the C-FROST interface \cite{hereid2018rapid} on a laptop with an Intel Core i7-6820 HQ CPU @ $2.7$ GHz with $16$ GB RAM, and consisted of $8418$ variables with $4502$ equality and $5880$ inequality constraints. Using each gait as an initial guess for the next speed in the library, the average number of iterations per run was $199$ with an average total evaluation time of $263.8$ seconds, and an average objective value of $\mathcal{J}(\mathbf{w}^*) = 4.12$.

\vspace{0.2cm}
\noindent\textbf{Extracting parameters for real-time control.} 
The controller implemented on hardware needs both the feedback control objectives, defined by $y^d(\alpha,t)$, and acceleration information $\ddot{q}^*$ from the optimal path to complete the planned motions. The feedback parameters, $\alpha$, already concisely parameterize the feedback control, and are placed in a large matrix which can be used in a bilinear interpolation routine. 
A subset of the resulting library of output parameters are shown in \figref{fig:opt_output_continuum_feedforward}, where the leg length, leg pitch, and hip roll outputs are visualized over various walking speeds alongside a corresponding motion in simulation. An interesting characteristic which can be immediately seen is that the stance leg length output has a `double-hump' shape, which is a result of the planned motions taking advantage of the spring deflection in the leg to minimize torque \eqref{eq:cost}.

Generalized accelerations $\ddot{q}^*$ are extracted directly from the optimization variables, $\mathbf{w}$, as time-series data from each step. To allow for easier implementation, regression is performed on each curve to obtain the parameters for a $6$th order B\'ezier polynomial. They can then be stacked with the $\alpha$ parameters in the same bilinear interpolation routine for code efficiency. Plots of the accelerations for the floating base $x$ and $y$ coordinates are visualized in \figref{fig:ddq_plots}. Finally, the floating-base position  $p^*_{x,y}$ and velocity $v^*_{x,y}$ relative to the stance foot is also extracted in the same manner as $\ddot{q}^*$. This is shown on the velocity plots in \figref{fig:main_experimental_plot} as the dashed lines and serves as a reference velocity for regulation.

\section{Real-time Tracking of Optimized Motions}
\label{sec:implementation}
The optimization framework described in Sec. \ref{sec:opt} creates a continuum of walking trajectories \cite{da2017supervised}, which can be interpolated to obtain walking motions at a range of speeds. In previous work we presented on Cassie \cite{reher2019cassie}, PD control achieved reasonable tracking of the desired outputs with the passive compliance in the system matching the planned response. However, implementation relied on high-gain PD control with heuristic feedforward terms and trajectory modifications that needed to be hand-tuned to account for the model mismatch. A similar approach was used also by others \cite{gong2018feedback}, which demonstrated walking without planning for compliance or velocity terms in the dynamics. A benefit to inverse dynamics approaches on robotic systems is that lower gain PD feedback control can be used, while feedforward terms which respect the constrained rigid body dynamics of the system are used to produce most of the control input.

\subsection{Inverse Dynamics Controller} \label{sec:controller}
The use of inverse dynamics control for underactuated and floating-base robots is significantly more complex than for fixed-base manipulators. In developing controllers for these systems, there are many considerations to address such as numerical problems due to repeated matrix inversions of the inertia matrix, contact force distribution, and computational efficiency \cite{reher2020inverse}. Here, we follow the approach of Mistry \cite{mistry2010decompprojection}, which uses an orthogonal decomposition to compute the inverse dynamics torques in the null-space of the constraints. 

The underlying idea of the method is to use a QR decomposition of the constraint Jacobian matrix:
\begin{align}
    J(q)^\mathsf{T} = Q \begin{bmatrix} R \\ 0 \end{bmatrix},
\end{align}
where $Q$ is an orthogonal matrix and $R$ is an upper triangle matrix of rank $m_\eta$. 
The $Q$ in the QR decomposition provides a coordinate transform which separates the system dynamics \eqref{eq:eom} into the constrained and unconstrained components:
\begin{align}
    Q^\mathsf{T} \left[ D(q) \ddot{q} + H(q,\dot{q}) \right] = Q^\mathsf{T} B u + \begin{bmatrix} R \\ 0 \end{bmatrix} \lambda. \label{eq:proj_dynamics}
\end{align}

\begin{remark}
    In \eqref{eq:proj_dynamics} the spring force term, $B_s \tau_s$, is not present. We originally developed the controller with the spring dynamics and were able to achieve walking. However, the robot would often overreact to perturbations in the spring deflections. Instead, 
    we apply 
    a constraint on the springs similar to \eqref{eq:swing_rigid_constraint}, but at their current deflected state. This produces a torque consistent with the deflected leg configuration, but does not react aggressively to spring oscillations. 
\end{remark}

Through the use of a selection matrix $S_u = \begin{bmatrix} 0_{(n-k)\times k} & I_{(n-k)\times(n-k)} \end{bmatrix}$, we can obtain the unconstrained system dynamics which are not dependent on the constraints:
\begin{align}
    S_u Q^\mathsf{T} \left[ D(q) \ddot{q} + H(q,\dot{q}) \right] = S_u Q^\mathsf{T} B u.
\end{align}
Solving for $u$ leads us to the inverse dynamics:
\begin{align}
    u_{ff} = (S_u Q^\mathsf{T} B)^{\dagger} S_u Q^\mathsf{T} \left[ D(q) \ddot{q}^d + H(q,\dot{q}) \right], \label{eq:inverse_dynamics}
\end{align}
where $\dagger$ represents the appropriate Moore-Penrose inverse and $\ddot{q}^d$ is a target acceleration for the system to follow. For a more detailed analysis of the orthogonal decomposition and controller derivation, we again direct readers to \cite{mistry2010decompprojection}. In addition to the feedforward term provided by \eqref{eq:inverse_dynamics}, a feedback correction is computed as a standard PD controller:
\begin{align}
\label{eq:torque}
    u = u_{ff} - \left(\frac{\partial y(t,q)}{\partial q_m}\right)^\dagger \left( K_p y(t, q) + K_d \dot{y}(t, q,\dot{q}) \right),
\end{align}
where $K_p=[900, 500, 300, 250, 200, 200, 200, 200, 25]$ and $K_d=[12, 6, 4, 6, 5, 6, 4, 4, 2]$ are the proportional and derivative gains applied to the output feedback on hardware.

In order to select the parameters which are used for output tracking and for the feedforward term, we use average velocity of the previous step, $\bar{v}^a_{k-1}$. Thus, our B\`ezier polynomial array provides us at each time instant with the nominal acceleration $\ddot{q}^*(t, \bar{v}^a_{k-1})$, outputs $y^d(t,\bar{v}^a_{k-1})$, floating-base positions $p^*_{x,y}(t, \bar{v}^a_{k-1})$ and velocities $v^*_{x,y}(t, \bar{v}^a_{k-1})$.

\begin{figure*}[t]
	\centering
	\vspace{2mm}
	\includegraphics[width= 1\textwidth]{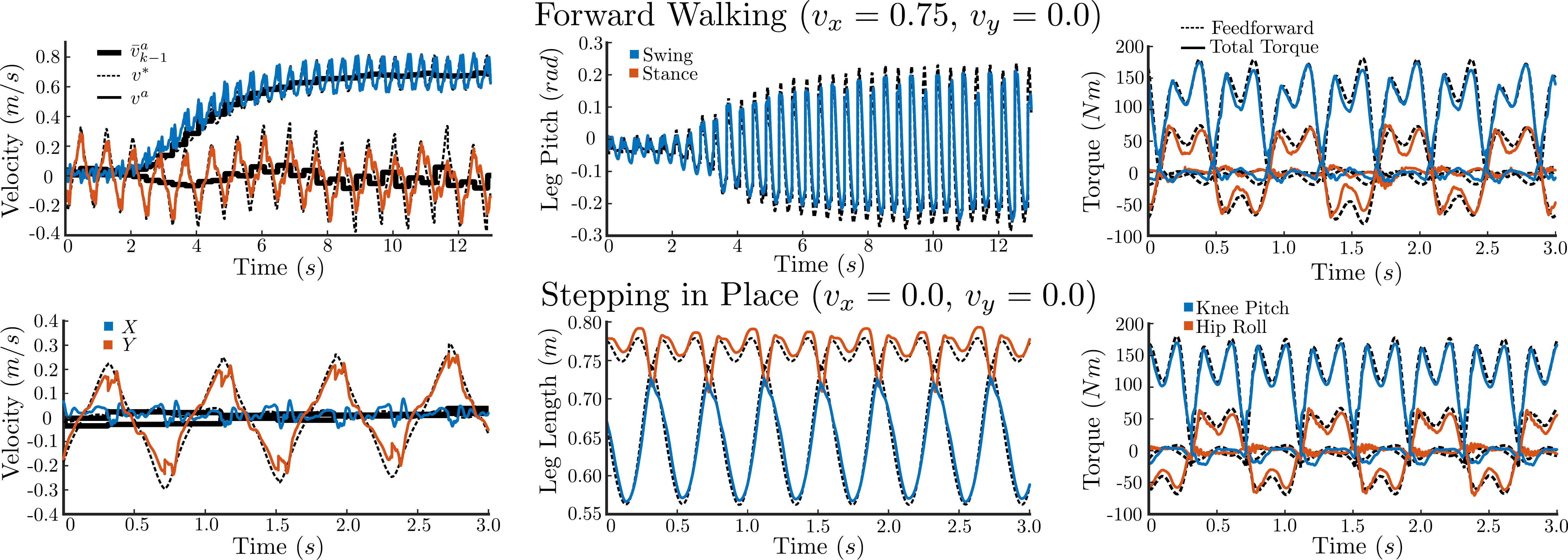}
	\caption{Experimental results for walking (top) and stepping in place (bottom). On the left is a comparison of the desired velocities from the current gait, $\bar{v}_{k-1}^a$, compared with the actual velocity of the robot. The center plots depict the accuracy of the tracking on the leg length and swing leg pitch outputs. On the right is a comparison of the torque from the feedforward controller with the total torque applied for the knee pitch and hip roll joints. }
	\label{fig:main_experimental_plot}
	\vspace{-5mm}
\end{figure*} 
%

\subsection{Velocity Tracking and Regulation Heuristics} \label{sec:heuristics}
The presented controller required no heuristic tuning on the feedforward terms to achieve tracking of the outputs such as offsets or hand-tuned torque profiles. However, directly implementing this controller with the trajectory obtained from the NLP \eqref{eq:opteqs} can at best result in a marginally stable locomotion for experiments as it has no way to trend walking towards an overall target walking speed. 
Motivated by this, a PD controller is used to find an offset to the footstrike location, and translated to the desired outputs:
\begin{align*}
    \Delta := \begin{bmatrix} \Delta_x \\ \Delta_y \\ \Delta_z \end{bmatrix} &= 
    \begin{bmatrix} \tilde{K}_{p,x} (\tilde{v}_{x}^{a} - v^d_x) + \tilde{K}_{d,x} (\tilde{v}_{x}^{a} - \bar{v}_{x,k-1}^a) \\
    \tilde{K}_{p,y} (\tilde{v}_{y}^{a} - v_y^d) + \tilde{K}_{d,y} (\tilde{v}_{y}^{a} - \bar{v}_{y,k-1}^a) \\
    0
 \end{bmatrix}, \\
    y_{sw,ll}^d &= || p_{nsf}^* + \Delta ||_2, \\
    y_{lp}^d &= \sin^{-1} \left( \frac{p_x^*(y^d) + \Delta_x}{y_{sw,ll}^d} \right) - y_{b,x}^d, \\
    y_{lr}^d &= \sin^{-1} \left( \frac{p_y^*(y^d) + \Delta_y}{y_{sw,ll}^d} \right) - y_{b,y}^d,
\end{align*}
where $\tilde{v}_{x,y}^a = \bar{v}_{k-1}^a + \left( v^a_{x,y} - v^*_{x,y} \right)$ is the current step velocity, $v^d_{x,y}$ is the target step velocity from the user joystick, $v^a_{x,y}$ is the instantaneous velocity of the robot relative to the stance foot, and $p_{nsf}^*(y^d) = ( p_{nsf,x}^*, p_{nsf,y}^*, p_{nsf,z}^* )$ are the nominal Cartesian swing foot positions computed from the desired outputs. This style of regulator is inspired by early work of \cite{raibert1984experiments}, and has been used widely in the literature.

An additional regulator is applied to modify the nominal accelerations of the floating base accelerations, which were pictured in \figref{fig:ddq_plots}. Through the application of a heuristic feedback controller, we can make the robot choose feedforward torques which trend the robot toward the target velocity while satisfying the contact constraints on the system:
\begin{align*}
    \ddot{q}^d_{x,y} = &\ddot{q}^*_{x,y} + k_p \left(p^a_{x,y}(q) - p^*_{x,y}\right) + \\ 
        &k_v (\tilde{v}_{x,y}^{a} - v^d_{x,y}) + k_i \int_0^t  \gamma (\tilde{v}_{x,y}^{a}(t') - v^d_{x,y}(t')) dt',
\end{align*}
where $k_p = [ 1.25, 1.90 ]$ is a gain affecting the $x$ and $y$ position errors of the pelvis relative to the stance foot, $k_v = [ 0.80, 0.60 ]$ is a gain on the step velocity tracking error, and $k_i = [ 1.90, 0.0 ]$ is a gain on the accumulated step velocity error with a decay constant of $\gamma = 0.9995$ to avoid integral windup. Because the feedforward term \eqref{eq:inverse_dynamics} uses the full actuator matrix $B$, it does apply some control effort on the stance foot actuator. It is for that reason that the integral term is applied to the $x$ direction, and not in the $y$.

\subsection{Experimental Implementation and Results}
The controller is implemented on the Intel NUC computer which comes installed in the Cassie torso, on which we added a PREEMPT\_RT kernel. 
The software runs on two ROS nodes: one which communicates to the Simulink Real-Time xPC over UDP to send torques and receive sensor data and to perform estimation, and a second which runs the controllers. Each node is given a separate core on the CPU, and is elevated to real-time priority. 
The first node runs at $2$ kHz and executes contact classification, inverse kinematics to obtain the heel spring deflection, and an EKF for velocity estimation \cite{bloesch2013state,reher2019cassie}.
The second node runs at $1$ kHz and receives the estimation and proprioceptive data over ROS. It then runs either the standing controller presented in \cite{reher2020inverse}, or executes the controllers presented in Sec. \ref{sec:controller} and \ref{sec:heuristics} before communicating the commanded torque over ROS. 

\begin{figure}[t]
	\centering
	\includegraphics[width= 0.98\columnwidth]{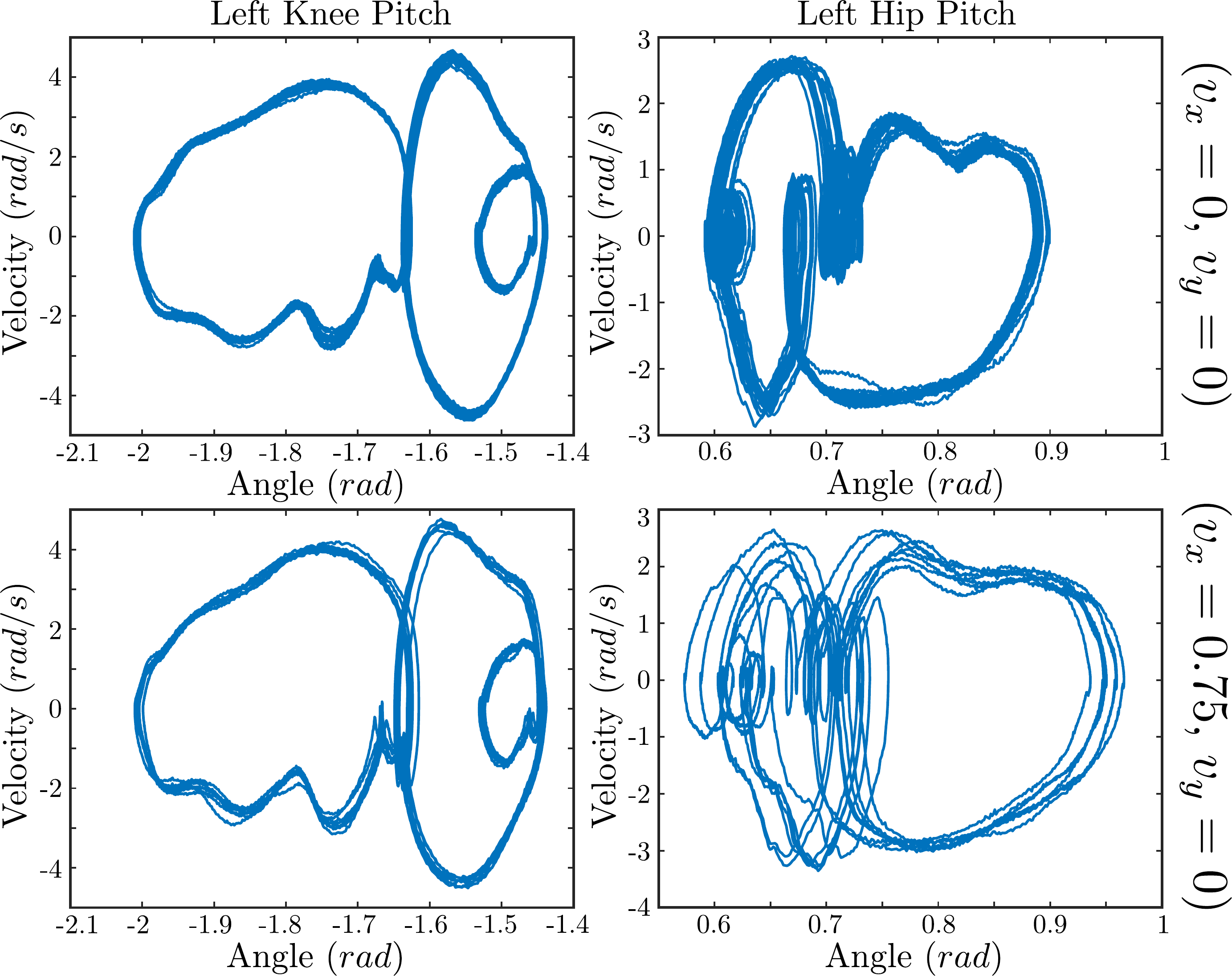}
	\caption{Phase portrait of the left knee and hip pitch joints for forward walking and stepping in place.}
	\label{fig:experiment_phase_portrait}
	\vspace{-6mm}
\end{figure}

\begin{figure*}[t]
	\centering
	\vspace{2mm}
	\includegraphics[width= 0.96\textwidth]{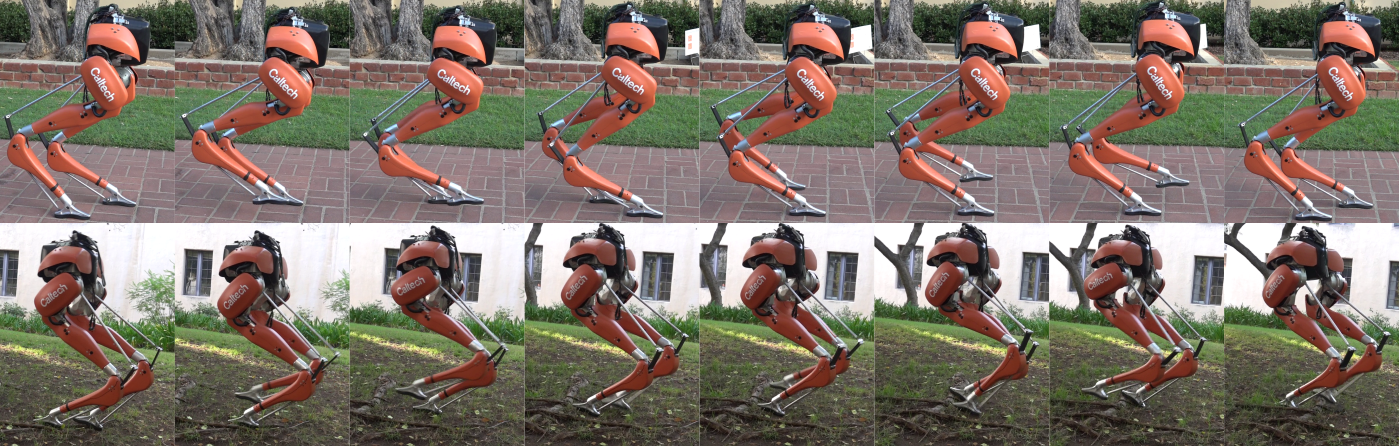}
	\caption{Gait tiles of the robotic walking exhibited experimentally on the Caltech Cassie robot. The top half is two steps of flat-ground forward walking outside, while the bottom half shows several steps of the robot walking over a height variation caused by both a shallow grassy slope and exposed roots. }
	\label{fig:gaittile}
	\vspace{-6mm}
\end{figure*}

These controllers were first tuned in a Gazebo simulation, shown in \figref{fig:opt_output_continuum_feedforward}, before implementation on hardware. Cassie was then tested indoors for stepping in place and forward, backward, lateral, and diagonal walking. The velocity tracking data, output tracking, and feed-forward torques are shown in \figref{fig:main_experimental_plot} for forward walking and stepping in place. Additionally, limit cycles of the knee and hip pitch joints are shown in \figref{fig:experiment_phase_portrait}, demonstrating the stability of the walking. Cassie was then taken outside to walk over several hills, raised roots, and a brick path. A video of both the simulation and experiments is provided at \cite{papervideo}, with walking tiles of the robot on flat ground and traversing rough terrain is shown in \figref{fig:gaittile}. The controller, simulation, and trajectory optimization  also made available in an open-source repository \cite{papergithub}.

%

\section{Conclusions and Future Work}
\label{sec:conclusions}
The Cassie biped poses a unique challenge due to its compliant leg mechanism and the highly underactuated nature of the dynamics. 
In order to leverage these components in experiments, we constructed a hybrid model for walking dynamics based on a compliant model. 
A trajectory optimization was then developed to efficiently generate walking trajectories using the method of HZD and was then tracked through an inverse dynamics controller. 
The resulting experiments show that the optimization can effectively capture the passive dynamics on a highly complex robot, while providing accurate model-based feedforward information.

On the physical system, a trivial double-support domain is virtually impossible to attain due to the compliance present in both legs. 
While this assumption simplified the development of controllers and trajectory planning, the consideration of a continuous double-support domain can enhance the stability of the behaviors. 
Future work will focus on reintroducing a nontrivial double-support domain to our walking model. 
Additionally, the use of an analytical solution to the inverse dynamics cannot account for optimal distributions of contact forces, friction, torque limits, or provide convergence guarantees. 
Thus, our future controller development is focused on further developing real-time model based controllers for implementation on hardware, such as extending the control Lyapunov function based methods in \cite{reher2020inverse} to walking.


\newpage
\bibliographystyle{plain}
\bibliography{bibdata}

\end{document}